\documentclass{article}

\usepackage[a4paper,margin=1in]{geometry}
\usepackage{authblk}
\usepackage{graphicx}
\usepackage{amsmath}
\usepackage{amsfonts}
\usepackage[numbers]{natbib}
\usepackage{hyperref}

\title{A Data-Free Symbolic Regression Approach for Solving Equations}

\author[1]{Sergei Garmaev}
\author[1]{Vinay Sharma}
\author[1]{Olga Fink}

\affil[1]{Intelligent Maintenance and Operations Systems Laboratory, EPFL, Lausanne, CH-1015, Switzerland}

\date{}

\begin{document}

\maketitle

\begin{abstract}
Many equations arising in science currently cannot be solved by available analytical techniques and are therefore solved numerically, without yielding explicit symbolic expressions. Existing symbolic regression approaches can recover symbolic expressions, but require training data obtained from the underlying process, rather than the governing equation alone. We propose the Symbolic Equation Solver (SES), a framework that formulates equation solving as an optimization problem over differentiable symbolic models. SES constructs its objective from the equation together with initial or boundary conditions, eliminating the need for paired input-output data. The learned model is expressed in explicit symbolic form, enabling further analysis. We evaluate SES on representative algebraic and differential equations, including a system of algebraic equations, an equation with transcendental terms, an ordinary differential equation, and partial differential equations with different initial or boundary conditions. Across these settings, SES recovers compact symbolic expressions that match the corresponding analytical solutions.
\end{abstract}

\section{Introduction} \label{sec:introduction}

Equations are fundamental to science and engineering as concise mathematical representations of relations between quantities. They formulate governing laws, constraints, and auxiliary conditions in a precise mathematical form. Solving an equation amounts to identifying an admissible object that satisfies all imposed relations simultaneously. Ideally, this is achieved analytically, yielding an explicit closed-form solution. However, such solutions are often unavailable. For instance, equations involving transcendental terms may not be solvable by standard analytical techniques, and for differential equations,  closed-form solutions are available only for specific classes of problems. As a result, most problems are addressed using numerical methods that produce approximate solutions. While being quantitatively accurate, these solutions typically provide limited insight into the underlying structure. By contrast, even approximate symbolic expressions can typically reveal functional dependencies, symmetries and qualitative behavior that remain difficult to extract from purely numerical solutions. 

Existing approaches to solving equations can be broadly divided according to the form of the resulting solution. Analytical methods and symbolic regression aim to recover explicit expressions, whereas numerical approximation methods, including residual-based neural approaches, seek to approximate solutions without recovering closed-form symbolic structure. Classical symbolic solvers derive exact solutions through analytical transformations, but their applicability is inherently limited by the structure of the equation \cite{winkler2024symbolic}. Symbolic regression instead aim to recover interpretable expressions from the data while balancing accuracy and simplicity \cite{makke2024interpretable}, using techniques such as genetic programming \cite{tonda2025review, burlacu2020operon}, reinforcement learning \cite{tian2025interactive}, or gradient-based optimization \cite{brunton2016discovering, liu2025kan}. Within differentiable symbolic regression,  methods vary substantially in expressivity.  Sparse Identification of Nonlinear Dynamics \cite{brunton2016discovering} identifies sparse linear combinations from predefined libraries of candidate terms, whereas  nonlinear symbolic models directly parameterize nonlinear symbolic structures. A representative example is the Equation Learner (EQL) \cite{sahoo2018learning}, which  replaces conventional neural activations  with symbolic operators, while sparsity is enforced in the connecting weights. This differentiable formulation enables gradient-based optimization and  integration into broader learning frameworks. However, existing symbolic regression methods, including EQL, are typically formulated from paired input-output observations, whereas many equation-solving problems are specified through governing equations together with initial or boundary conditions rather than supervised datasets.

Residual-based neural methods address this formulation more directly by optimizing a model to satisfy the equations and auxiliary constraints at collocation points. This idea appeared in early neural network approaches to differential equations \cite{lagaris1998artificial} and was later developed further in Physics-Informed Neural Networks (PINNs) \cite{raissi2019physics}. These methods are well suited to equation-based problem settings, but they produce numerical approximations encoded in neural networks rather than explicit symbolic expressions. As a result, the recovered solution is not in a form suitable for further analysis.

Analytical methods therefore recover explicit solutions only for equations whose structure admits tractable derivation, while symbolic regression methods require supervised input-output observations. Residual-based neural approaches, in contrast, operate directly from governing equations and auxiliary conditions but return neural surrogates rather than symbolic expressions. Existing approaches therefore fail to jointly satisfy three desirable properties: (i) operating directly from governing equations and auxiliary constraints, (ii) avoiding supervised input-output data, and (iii) recovering the solution in explicit symbolic form.
This leaves a clear gap between equation-based formulations and symbolic recovery of the solution.

In this work, we propose Symbolic Equation Solver (SES), a framework that formulates equation solving as optimization over differentiable symbolic models. SES treats the symbolic expression itself as the optimization object. Unlike traditional symbolic regression methods, including EQL, which learn expressions by fitting paired input-output data, SES uses the governing equation and auxiliary conditions directly as the training signal. The symbolic network represents a candidate solution, and its parameters are optimized to minimize equation and constraint residuals at sampled collocation points. After optimization, the trained symbolic network is converted into an explicit expression. In this way, SES searches for a symbolic solution consistent with the given equation and auxiliary conditions without requiring paired target data.

Unlike conventional numerical solvers and residual-based neural approaches such as PINNs, which typically return numerical approximations of the solution, SES is designed to recover the solution in explicit symbolic form. This is achieved by optimizing a differentiable symbolic model against residual terms derived from the governing equation together with any initial or boundary conditions, which allows the same formulation to be applied to both non-differential and differential equations without supervised input-output data. The framework further extends naturally from a single unknown quantity to systems with multiple unknowns optimized jointly under a shared objective. SES formulates symbolic solution recovery as equation-constrained optimization over differentiable symbolic models. Across the equations studied in this work, this leads to compact symbolic expressions that match the corresponding analytical solutions.

\section{Related Work} \label{sec:related_work}
Symbolic approaches to equation solving manipulate mathematical expressions exactly in order to obtain explicit symbolic solutions. In computer algebra systems, this includes operations such as simplification, factorization, elimination, differentiation, integration, and equation specific transformations applied to mathematical objects \cite{von2003modern}. For polynomial systems, Gröbner-basis methods provide a systematic algebraic tool for elimination and exact solution analysis \cite{cox2025ideals}. For algebraic differential equations, symbolic solution methods can also rely on algebraic-geometric constructions, in which symbolic solutions are obtained through parameterizations and elimination procedures \cite{winkler2024symbolic}. For ordinary differential equations, exact symbolic methods are available only for restricted classes. For example, Kovacic's algorithm finds closed-form Liouvillian solutions of second-order linear homogeneous differential equations with rational-function coefficients when such solutions exist \cite{kovacic1986algorithm}. Although these methods can recover exact symbolic solutions, their applicability is fundamentally constrained by the functional structure of the underlying equation class. Consequently, they do not provide a general mechanism for recovering symbolic solutions to arbitrary nonlinear, transcendental, or differential equations.

Symbolic regression instead provides a data-driven route to explicit mathematical expressions by searching for formulas that fit observed input-output samples while balancing accuracy and complexity \cite{makke2024interpretable}. Modern approaches differ mainly in how the expression space is explored. Evolutionary methods represent candidate formulas as a sequence of expression tree nodes and optimize them through genetic operations \cite{koza1994genetic}. Recent implementations, such as Operon \cite{burlacu2020operon} and PySR \cite{tonda2025review}, improve this paradigm through efficient encodings, scalable search, and constant optimization. Beyond evolutionary search, deep learning-based symbolic regression methods use neural models to predict the expression or its functional structure from the input data points. Deep Symbolic Regression trains an autoregressive recurrent neural network to generate mathematical expressions using a risk-seeking policy-gradient objective \cite{petersen2021deep}, while recent reinforcement-learning-based extensions introduce interactive or offline reinforcement learning mechanisms to incorporate user feedback and guide the symbolic search process \cite{tian2025interactive}. Neural Symbolic Regression that Scales pretrains a transformer model on procedurally generated equations and predicts expressions from input-output points, using the model output to guide equation search \cite{biggio2021neural}. End-to-end transformer approaches further train sequence models to predict full symbolic expressions, including numerical constants, directly from observed function values \cite{kamienny2022end}. Despite substantial progress in symbolic recovery from observations, these methods remain fundamentally supervised: they assume access to input-output samples of the target relation rather than the governing equation itself.

Differentiable symbolic models take a different route by embedding symbolic structure directly into trainable architectures. Sparse Identification of Nonlinear Dynamics restricts the search to sparse linear combinations of predefined candidate functions and identify governing equations from measurement data \cite{brunton2016discovering}. Kolmogorov-Arnold Networks (KANs) replace scalar weights with learnable univariate functions on edges, which can be interpreted or simplified after training to obtain symbolic expressions \cite{liu2025kan}. However, KANs do not impose exact symbolic operations at the nodes during training. Symbolic structure instead is obtained by interpreting, simplifying, or fitting the learned univariate functions after optimization. In contrast, EQL replaces standard neural activations with predefined symbolic operators and learns concise equations through gradient-based optimization \cite{martius2016extrapolation}. Several extensions of the EQL framework have expanded its expressivity and training strategy: EQL$^{\div}$ introduces division units with a stabilizing curriculum \cite{sahoo2018learning}, Informed Equation Learning incorporates expert knowledge, structured sparsity, and more robust training for singular atomic units \cite{werner2021informed}, and Complex Equation Learner extends EQL by optimizing complex-valued weights and projecting the output back to the real domain, which allows gradient-based training to recover rational and singular expressions involving division, logarithms, and square roots \cite{garmaev2026ceql}. In this work, we adopt the original EQL formulation as the symbolic model class but  change the learning setting itself: instead of supervised regression from data, we optimize symbolic models directly against equation residuals and auxiliary constraints.

Residual-based neural methods address equation-defined problems by training neural networks to satisfy differential equations and auxiliary constraints. Early work \cite{lagaris1998artificial} represented trial solutions using feed-forward neural networks. The trial form was constructed to satisfy initial or boundary conditions, while the network parameters were optimized to satisfy the differential equation. Physics-Informed Neural Networks later generalized this residual minimization perspective by incorporating differential equation residuals, initial and boundary conditions, and, observational data into a unified training loss function \cite{raissi2019physics}. Related mesh-free neural solvers, such as the Deep Galerkin Method, approximate PDE solutions with neural networks trained to satisfy the differential operator, initial conditions, and boundary conditions at randomly sampled space-time points \cite{sirignano2018dgm}. Variational approaches such as the Deep Ritz method insted train a neural networks by minimizing energy functionals associated with variational problems, particularly those arising from PDEs \cite{yu2018deep}. These methods are well aligned with equation-based problem formulations and can avoid supervised solution data, but the learned solution remains encoded in neural network parameters. SES follows the same equation-driven optimization principle, but replaces the neural function approximator with a differentiable symbolic model, enabling the optimized solution to be recovered  as an explicit expression.

\section{Methods} \label{sec:methods}
The proposed SES formulates equation solving as optimization in a  differentiable symbolic function space. Given a governing equation and, where applicable, auxiliary constraints such as initial or boundary conditions, SES searches for a symbolic expression that minimizes a residual-based objective derived directly from these relations. In contrast to conventional symbolic regression, SES does not require paired input–output observations; the equation itself  defines the supervision signal. Candidate solutions are represented by differentiable symbolic models whose parameters are optimized through gradient-based learning. Once optimized, the learned symbolic structure can be extracted and verified independently.

More generally, SES considers problems specified by one or more governing relations
\begin{equation}
    \mathcal{F}_k\big(\mathbf{x}, \mathbf{u}(\mathbf{x}), \nabla \mathbf{u}(\mathbf{x}), \nabla^2 \mathbf{u}(\mathbf{x}), \ldots \big)=0,
    \qquad k=1,\ldots,K,
\end{equation}
defined on a domain $\Omega$, where $\mathbf{x}\in\Omega$ denotes the independent variables and $\mathbf{u}:\Omega\to\mathbb{R}^m$ denotes the unknown scalar- or vector-valued function. For differential equations, the operators $\mathcal{F}_k$ may depend on derivatives of $\mathbf{u}$ with respect to the components of $\mathbf{x}$. When initial or boundary conditions are present, they are written as additional residual relations
\begin{equation}
    \mathcal{B}_j\big(\mathbf{x}, \mathbf{u}(\mathbf{x}), \nabla \mathbf{u}(\mathbf{x}), \ldots \big)=0,
    \qquad j=1,\ldots,J,
\end{equation}
imposed on the corresponding initial-time set or boundary subset. The objective is to find an explicit symbolic expression, or a collection of expressions, $\hat{\mathbf{u}}(\mathbf{x})$, that satisfies the governing relations and all imposed auxiliary conditions.

\begin{figure}[t]
    \centering
    \includegraphics[width=\linewidth]{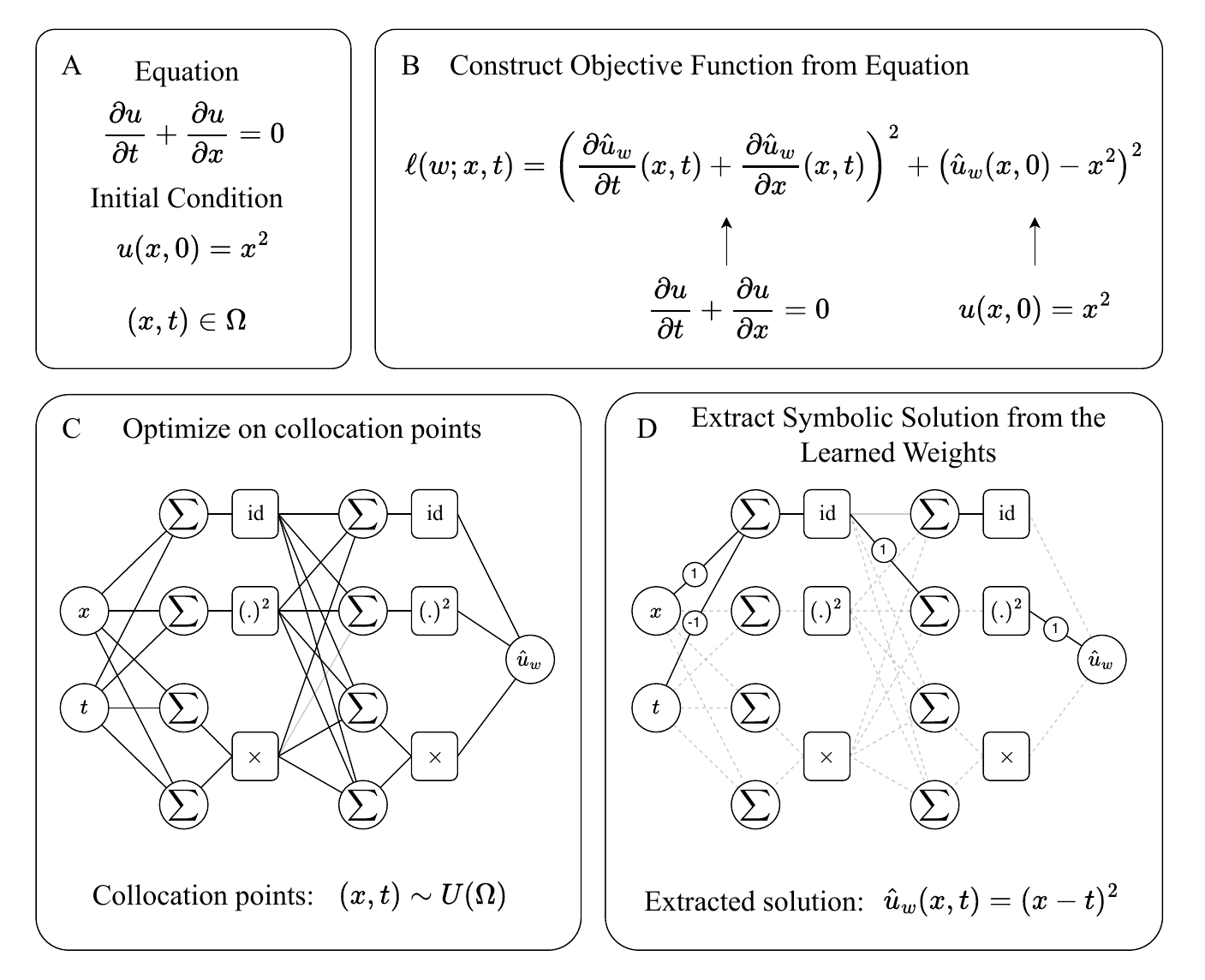}
    \caption{Schematic overview of the Symbolic Equation Solver (SES) framework. Given a governing equation and auxiliary conditions (A), SES constructs an objective function from the corresponding residual terms (B) and optimizes a differentiable symbolic model at collocation points (C). The resulting learned weights define a sparse symbolic network, from which the an explicit expression approximating the solution is extracted (D).}
    \label{fig:ses_illustration}
\end{figure}

The unknown quantity is represented by a differentiable symbolic model based on an EQL-type network
\begin{equation}
    \hat{\mathbf{u}}(\mathbf{x};\theta)
    =
    \left(\hat{u}_1(\mathbf{x};\theta_1),\ldots,\hat{u}_m(\mathbf{x};\theta_m)\right),
\end{equation}
where $\theta$ denotes the trainable coefficients of the symbolic network. For scalar problems, $m=1$, while for systems each component $\hat{u}_i$ represents one unknown. In the present implementation, each component is represented by an EQL-type symbolic network, implemented as a fully connected architecture in which conventional activation functions are replaced by symbolic operators, as illustrated in Figure~\ref{fig:ses_illustration}. This representation enables gradients to propagate through symbolic operators directly, allowing symbolic expressions to be optimized using standard first-order optimization methods.

The symbolic operators form a library of unary and binary functions, depending on whether they act on one or two inputs. A unary operator acts on a single weighted sum of activations from the previous layer, whereas a binary operator acts on two separate weighted sums. In this way, the network weights can be interpreted as coefficients of the resulting symbolic expression. In all experiments, the symbolic model has the same overall architecture. The maximum tree depth is set to two, corresponding to a two-layer fully connected network. Each layer contains the operator set $\{\text{id}, \text{const}, (.)^2, \exp(.), \tanh(.), \times\}$, and each operator is instantiated twice per layer. This operator library balances expressivity and optimization stability while remaining sufficiently compact to encourage sparse symbolic recovery. As illustrated in Figure~\ref{fig:ses_illustration}, the learned weights are used to extract an explicit symbolic expression by propagating the input variables through the learned coefficients and operators across the network layers.

The symbolic model is trained by minimizing an objective function constructed from the governing equation together with initial or boundary conditions when applicable. Specifically, all terms of the equation are moved to the left-hand side to define residuals $\mathcal{F}_k$, while initial and boundary conditions are written as auxiliary residuals $\mathcal{B}_j$. These residuals are functions of the symbolic model, and for differential equations may also depend on its derivatives with respect to the independent variables. Therefore, the objective is evaluated at sampled points in the corresponding domains. We refer to these points as collocation points.

Let $\{\mathbf{x}_i\}_{i=1}^{N_F}$ denote collocation points in the domain of the governing equation, and let $\{\mathbf{z}_i\}_{i=1}^{N_B}$ denote points on the corresponding boundary or initial-time set where auxiliary conditions are imposed. SES minimizes
\begin{equation}
\begin{aligned}
    \mathcal{L}(\theta)
    &=
    \frac{1}{N_F}\sum_{i=1}^{N_F}
    \sum_{k=1}^{K}
    \left|
    \mathcal{F}_k\big(
        \mathbf{x}_i,
        \hat{\mathbf{u}}(\mathbf{x}_i;\theta),
        \nabla \hat{\mathbf{u}}(\mathbf{x}_i;\theta),
        \nabla^2 \hat{\mathbf{u}}(\mathbf{x}_i;\theta),
        \ldots
    \big)
    \right|^2 \\
    &\quad+
    \frac{1}{N_B}\sum_{i=1}^{N_B}
    \sum_{j=1}^{J}
    \left|
    \mathcal{B}_j\big(
        \mathbf{z}_i,
        \hat{\mathbf{u}}(\mathbf{z}_i;\theta),
        \nabla \hat{\mathbf{u}}(\mathbf{z}_i;\theta),
        \ldots
    \big)
    \right|^2 .
\end{aligned}
\label{eq:ses_loss}
\end{equation}
Here, $K$ is the number of governing residuals, $J$ is the number of auxiliary residuals, $N_F$ is the number of governing-equation collocation points, and $N_B$ is the number of auxiliary-condition collocation points. When no auxiliary conditions are present, the second term in Eq.~\eqref{eq:ses_loss} is omitted. Derivatives of $\hat{\mathbf{u}}$ appearing in the residuals are computed by automatic differentiation.

Under exact optimization and sufficiently expressive symbolic representations, any model attaining zero loss satisfies all imposed equations and auxiliary constraints exactly. In practice, however, the optimization is carried out over the symbolic function class induced by the SES architecture and evaluated at finitely many collocation of points. Therefore, the minimization procedure generally produces an approximate symbolic solution candidate, while exact validity must be established separately by analytical substitution or by explicit verification that the recovered expression satisfies the governing equation and all auxiliary conditions.

SES  does not require a supervised training dataset, but collocation points are needed to evaluate the objective function, since the model values and, when required, their derivatives must be computed at specific inputs. For non-differential equations, the solution is given by constants, which can be viewed as constant functions of an auxiliary variable. In such cases, SES represents the unknown quantity as a function of this auxiliary variable, and sampled values of the auxiliary variable serve as collocation points. For differential equations, where the solution is naturally a function of one or more coordinates, the auxiliary variable is replaced by the corresponding spatial or spatio-temporal input variables. The objective function is then evaluated at collocation points sampled from the domain of interest. For each experiment, 256 collocation points are drawn independently at each training epoch.

The optimization process is performed using Adam optimizer \cite{kingma2015adam} and proceeds in three phases. In the first phase, the model is trained for $50,000$ epochs using only the residual objective. In the second phase, training continues for $200,000$ epochs with an additional $\ell_1$-penalty on the internal weights to promote sparsity. During this phase, $25\%$ of the weights with the lowest contribution to the residual loss on the current training batch are periodically pruned. Pruning is repeated until a minimum number of active weights is reached, which is set to five in all experiments except for the Poisson equation, where a larger number of active weights is retained to represent the higher complexity of the solution. In the third phase, the $\ell_1$-penalty is removed and the remaining weights are fine-tuned for $20,000$ additional epochs using only the residual objective. The final symbolic solution is then extracted from the trained model after rounding the learned weights to five decimal places.

\section{Results} \label{sec:results}

We evaluate SES on algebraic, transcendental, ordinary differential, and partial differential equations to assess whether equation-constrained symbolic optimization can recover explicit symbolic solutions across diverse classes of equations. The experiments are designed to test three properties simultaneously: (i) recovery of exact or near-exact symbolic structure, (ii) applicability across both algebraic and differential settings, and (iii) optimization without supervised input-output targets. We begin with a system of linear equations and a transcendental equation that does not admit a solution through standard symbolic manipulations, such as algebraic rearrangement, substitution, or isolation of the unknown using elementary inverse functions. We then consider differential equations of increasing complexity, including a nonlinear ordinary differential equation, a first-order transport equation, and a two-dimensional Poisson equation with mixed boundary conditions. For these equations, the analytical solutions can be written using symbolic operations included in the model's symbolic library. The symbolic operation library is fixed for all equations and is described in detail in Section~\ref{sec:methods}. Across all examples, SES is trained solely from equation residuals and auxiliary constraints without supervised input-output data, and the recovered symbolic expressions are compared against the corresponding analytical solutions.

\subsection{Linear and Transcendental Equations} \label{sec:linear_and_transcendental}
We begin with the system of two linear equations
\begin{equation} \label{eq:linear_system}
    \begin{cases}
        2x + 3y = 7,\\
        x - y = 1.
    \end{cases}
\end{equation}
whose exact solution is $x = 2$ and $y = 1$. To recover this solution, we represent the two unknowns by separate SES models, $\hat{x}(t)$ and $\hat{y}(t)$, where $t$ is an auxiliary input sampled uniformly from $U(-2, 2)$. The auxiliary input is required here to evaluate the objective at different collocation points. To extract the solution, the symbolic computational graph retained after optimization and pruning is traversed layer by layer, composing the symbolic operators into an explicit expression. After optimization, the extracted symbolic expressions, rounded to five decimals, are $\hat{x}(t) = 2.0$, $\hat{y}(t) = 1.0$, which exactly reproduce the analytical solution. This experiment demonstrates that SES can recover constant symbolic solutions jointly for multiple unknown variables within a shared residual optimization problem.

We next consider the transcendental equation
\begin{equation} \label{eq:transcendental}
    e^x + x^3 = e + 1,
\end{equation}
which has the unique real solution $x = 1$. Since the unknown cannot be isolated by standard symbolic manipulations and this equation is typically handled numerically. Using a univariate SES model $\hat{x}(t)$, with the auxiliary variable $t$ sampled uniformly from $U(-2, 2)$, SES recovers the symbolic solution $\hat{x}(t) = 1.0$, matching the exact solution up to five decimal places. This result shows that SES can recover explicit symbolic solutions even when the equation itself does not admit straightforward analytical isolation of the unknown.

\subsection{Differential Equations} \label{sec:differential}
We next consider differential equations, where the unknown, in contrast to the previous cases, is a function. Compared to algebraic problems, the problems involving differential terms are more challenging and often admit only numerical solutions. Here, SES constructs the objective directly from residuals of the differential equations together with the corresponding initial or boundary conditions, while derivatives of the symbolic model are computed through automatic differentiation.

We first consider the nonlinear initial value problem
\begin{equation} \label{eq:ode}
    \frac{dy}{dt} = 1 - y^2, \qquad y(0)=0.
\end{equation}
with the exact solution $y(t) = \tanh t$. In this case, the auxiliary inputs of the SES model fall into natural place and coincides with temporal coordinate $t$. Here the SES model is represented by $\hat{y}(t)$. Derivative terms in the equation are evaluated through automatic differentiation. The objective function for this problem is constructed from the squared residuals of both the differential equation and the initial condition. The recovered symbolic expression, $\hat{y}(t) = 1.0 \tanh(1.0 t)$, matches the analytical solution with high precision. Importantly, the recovered expression preserves the correct nonlinear functional form rather than merely approximating the trajectory numerically.

We next consider the linear transport equation with quadratic initial condition
\begin{equation} \label{eq:heat_equation}
    \frac{\partial u}{\partial t} + \frac{\partial u}{\partial x} = 0, \qquad u(x,0)=x^2.
\end{equation}
Its exact solution is given by $u(x,t)=(x-t)^2$. The SES model $\hat{u}(x, t)$ is trained by minimizing residuals of the transport equation together with the residual of the initial condition, using collocation points sampled from $x \sim U(-2, 2)$ and $t \sim U(0, 2)$. SES recovers the symbolic solution
\begin{equation}
    \hat{u}(x, t) = 0.95184 (-1.025\,t + 1.025\,x)^2,
\end{equation}
which closely reproduces the analytical solution and recovers the correct underlying symbolic structure. Although the recovered coefficients are approximate, the learned expression remains symbolically equivalent to the true quadratic transport solution up to scaling and numerical rounding.
 
We further consider the two-dimensional Poisson equation
\begin{equation} \label{eq:poisson}
    \frac{\partial^2 u}{\partial x^2} + \frac{\partial^2 u}{\partial y^2} = 4 + 2x, \qquad (x,y)\in(0,1)^2,
\end{equation}
with mixed boundary conditions
\begin{equation} \label{eq:poisson_bc}
    u(0,y)=y^2, \qquad u(x,0)=x^2, \qquad \frac{\partial u}{\partial x}(1,y)=2+y^2, \qquad \frac{\partial u}{\partial y}(x,1)=2+2x.
\end{equation}
The exact solution is $u(x,y) = x^2 + y^2 + x y^2$. The choice of the boundary conditions is motivated by the set of symbolic operators available in the library specified in Section~\ref{sec:methods}. The SES model is optimized using collocation points sampled uniformly from the unit square, with the loss constructed from both PDE and boundary condition residuals. SES recovers
\begin{equation}
    \hat{u}(x,y) = 1.0\,x^2 + 1.0\,x\,y^2 + 1.0\,y^2
\end{equation}
which matches the analytical solution up to numerical precision. This experiment demonstrates that SES can recover multivariate symbolic structure involving coupled polynomial interactions under mixed boundary constraints.

We next consider the same Poisson equation under a different set of mixed boundary conditions. Since the Poisson equation alone does not uniquely determine the solution, this separates the contribution of the boundary constraints residuals from that of the interior PDE residual. The problem now is given by
\begin{equation}
    u(0,y)=0, \qquad u(x,0)=2x^2, \qquad \frac{\partial u}{\partial x}(1,y)=4+y^2, \qquad \frac{\partial u}{\partial y}(x,1)=2x.
\end{equation}
The corresponding exact solution is   $u(x,y) = 2x^2 + x y^2$. In this case, SES recovers $\hat{u}(x, y) = 2.0\,x^2 + 1.0\,x\,y^2$, again reproducing the exact symbolic solution.

Overall, these results show that SES can recover compact symbolic solutions across diverse classes of equations using a unified residual-based formulation, without requiring supervised target values. Across all settings, symbolic structure emerges directly from equation-constrained optimization rather than from regression against observed solution trajectories.

\section{Discussion} \label{sec:discussion}
The results  suggest that equation solving can be framed not only as analytical derivation or numerical approximation, but also as optimization in symbolic function space, where candidate expressions are constrained by equations and auxiliary conditions. Across the considered algebraic and differential problems, SES recovers compact symbolic solutions solely from residual minimization. This indicates that explicit solution discovery can emerge from equation-constrained optimization without supervised input-output data.

Conceptually, SES occupies a distinct position between symbolic regression and residual-based scientific machine learning. Unlike conventional symbolic regression, which is typically data-driven and seeks governing laws or functional relations from observations, SES targets symbolic solutions to prescribed equations without labeled input–output pairs. Conversely, residual-based neural solvers such as physics-informed neural networks optimize equation residuals but encode solutions implicitly in neural network parameters rather than explicit symbolic structure. SES instead formulates symbolic solution recovery itself as a differentiable optimization problem over symbolic representations. In this sense, the framework combines  equation-constrained learning with the interpretability, compactness, and analytical accessibility of symbolic models.

Recovering solutions in symbolic form is important not only for interpretability, but also because explicit expressions  expose structural dependencies and functional relationships, that remain difficult to infer from purely numerical approximations. Even approximate symbolic solutions can reveal candidate analytical forms, invariances, or simplifying structure that may support further theoretical analysis. More broadly, this suggests that symbolic solution recovery could serve not only as an  computational tool, but also as a mechanism for scientific hypothesis generation and discovery.

The experiments further show that the framework  naturally extends beyond single-variable settings, as illustrated by the joint recovery of multiple unknown quantities under shared residual constraints. This points toward a broader class of equation-constrained symbolic learning problems involving coupled algebraic and differential relations.

\section{Conclusion} \label{sec:conclusion}
This work introduces SES, a framework for recovering approximate symbolic solutions to equations directly from governing equations and auxiliary constraints without supervised input-output data. By optimizing differentiable symbolic models and optimizing them over equation and constraint residuals, SES combines the equation-driven formulation of residual-based neural solvers with symbolic regression. The results demonstrate feasibility on problems with known analytical solutions and relatively concise symbolic structure, including algebraic and differential equations with different boundary conditions.

Beyond these proof-of-concept examples, symbolic equation solving could be of use in settings where compact symbolic approximations are valuable, such as scientific computing, reduced order modelling, and control. The framework may also be relevant to AI-assisted mathematical discovery, particularly when symbolic structure and interpretability matter alongside predictive accuracy. Extending symbolic equation solving to more challenging problems will likely require more expressive symbolic architectures, broader operator libraries, and more robust optimization and pruning strategies. More broadly, the results suggest a route toward integrating symbolic reasoning more directly into scientific machine learning.

Several important directions remain for future work. Extending SES to more challenging PDEs, high-dimensional systems, and chaotic or stiff dynamics will require more expressive symbolic architectures, richer operator libraries, and more robust optimization strategies. Improving scalability and search efficiency is likely to require better sparsity induction, adaptive operator selection, and hierarchical symbolic representations. Another important direction is integrating symbolic priors, dimensional constraints, or physical invariances directly into the optimization process to guide symbolic discovery. More broadly, the results suggest that differentiable symbolic representations could provide a foundation for more interpretable and structurally grounded scientific machine learning systems.

\bibliographystyle{plainnat}
\bibliography{references}

\end{document}